\documentclass[10pt,twocolumn,letterpaper]{article}

\pdfoutput=1

\usepackage{times}
\usepackage{epsfig}
\usepackage{graphicx}
\usepackage{amsmath}
\usepackage{amssymb}

\usepackage[font=small,skip=0pt]{caption}

\usepackage{array}
\usepackage{mdwmath}
\usepackage{mdwtab}
\usepackage{url}
\usepackage{color}
\usepackage[]{graphicx}
\usepackage{epsfig}
\usepackage{bbm}
\usepackage{enumerate}
\usepackage{amsfonts}
\usepackage{amsmath}
\usepackage{amsthm}
\usepackage{amssymb}
\usepackage{cite}
\usepackage{algorithm}
\usepackage{algpseudocode}
\usepackage{eqparbox}
\usepackage{comment}
\usepackage[tight,footnotesize]{subfigure}
\usepackage{float}
\usepackage{mathtools}
\usepackage{centernot}
\usepackage{gensymb}
\usepackage{algpseudocode}
\usepackage{caption}
\usepackage{lipsum}
\usepackage{comment}

\DeclareMathOperator*{\argmin}{argmin}

\renewcommand{\algorithmiccomment}[1]{\hfill \eqparbox{COMMENT}{// #1}}

\usepackage[utf8]{inputenc}
\usepackage[english]{babel}

\begin{document}

\title{Multi-Resolution Data Fusion for Super-Resolution Electron Microscopy}

\author{Suhas Sreehari$^{1,*}$, S. V. Venkatakrishnan$^2$, Katherine L. Bouman$^3$, Jeffrey P. Simmons$^4$,\\ Lawrence F. Drummy$^4$, and Charles A. Bouman$^1$\\
{\small $^1$School of Electrical and Computer Engineering, Purdue University, West Lafayette, IN}\\
{\small $^2$ISML, Oak Ridge National Laboratory, Oak Ridge, TN}\\
{\small $^3$CSAIL, Massachusetts Institute of Technology, Cambridge, MA}\\
{\small $^4$Air Force Research Laboratory, Dayton, OH}\\
{\tt\small $^*$ ssreehar@purdue.edu}
}

\date{}

\twocolumn[{%
\renewcommand\twocolumn[1][]{#1}%
\maketitle
\begin{center}
    \centering
    \includegraphics[width=1\textwidth]{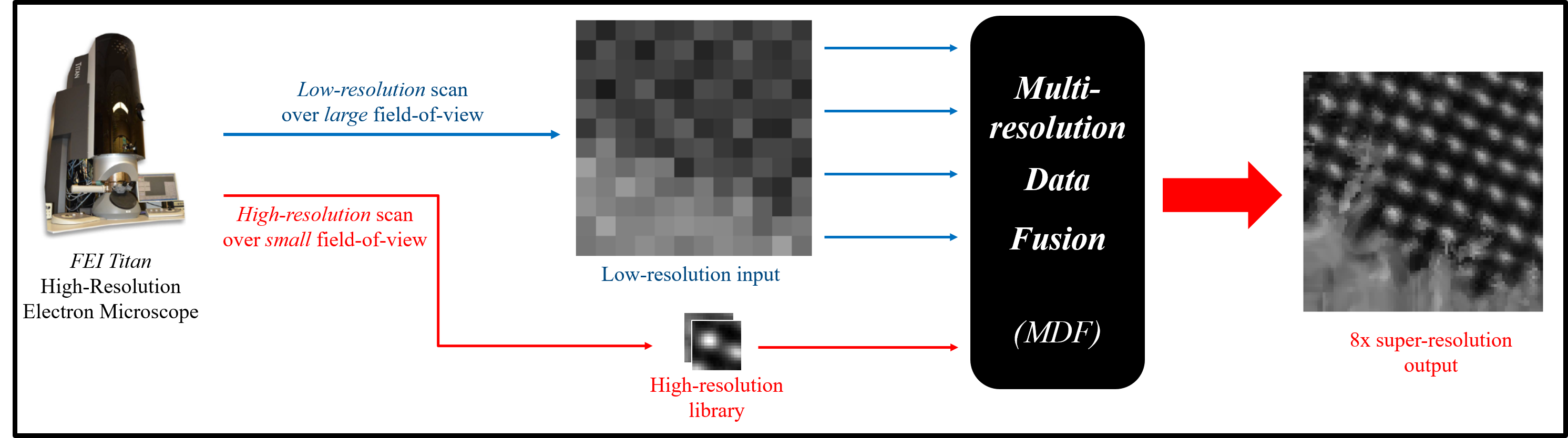}
    \captionof{figure}{
An illustration of the proposed multi-resolution data fusion (MDF) system
in which a transmission electron microscope is specially configured to collect both low resolution data over a large field-of-view (FoV)
together with a small set of high resolution patches from the same sample.
The MDF algorithm uses the high-resolution patches as an image model within the ``plug-and-play'' framework
to synthesize the 8x interpolated super-resolution output image over the full FoV.
We demonstrate the MDF system using an aberration-corrected FEI Titan transmission electron microscope
and show that it can dramatically speed up image acquisition and reduce dosage as compared to conventional homogeneous raster-scanning microscopy.
}
\end{center}%
}]
\nopagebreak

\begin{abstract}
\vspace{-3mm}
{\small \textit{Perhaps surprisingly, the total electron microscopy (EM) data collected to date is less than a cubic millimeter.
Consequently, there is an enormous demand in the materials and biological sciences 
to image at greater speed and lower dosage, while maintaining resolution.
Traditional EM imaging based on homogeneous raster-order scanning severely limits the volume 
of high-resolution data that can be collected, 
and presents a fundamental limitation to understanding physical processes such as material deformation, crack propagation, and pyrolysis.}}

{\small \textit{We introduce a novel multi-resolution data fusion (MDF) method for super-resolution computational EM.
Our method combines innovative data acquisition with novel algorithmic techniques 
to dramatically improve the resolution/volume/speed trade-off. 
The key to our approach is to collect the entire sample at low resolution,
while simultaneously collecting a small fraction of data at high resolution.
The high-resolution measurements are then used to create a material-specific patch-library that is used 
within the ``plug-and-play'' framework to dramatically improve super-resolution of the low-resolution data.
We present results using FEI electron microscope data that demonstrate 
super-resolution factors of 4x, 8x, and 16x, 
while substantially maintaining high image quality and reducing dosage.}}
\end{abstract}

\section{Introduction}

Scanning transmission electron microscopes are widely used for characterization of samples at the nano-meter scale. 
However, raster scanning an electron beam across a large field of view is time consuming and can often damage the sample. 
In fact, the sum total of all electron microscopy (EM) data collected to date, represents less than a cubic millimeter of material \cite{williams1996transmission}. 
This fact encapsulates a central challenge of materials and biological sciences to image larger volumes at higher resolution, greater speed, and lower dosage. 

Traditional microscopy based on homogeneous raster-order single-resolution scanning severely limits the volume of high resolution data that can be collected, and represents a fundamental limitation in understanding physical processes such as material deformation, crack propagation, and pyrolysis. 
For these reasons, there is growing interest in reconstructing full-resolution images 
from sparsely-sampled \cite{inpaintReviewSPM14, hyrumEI13, carinSparseInterp14} and low-resolution images \cite{farsiu2004fast, park2003super, glasner2009super, shechtman2005space}.

In many cases, EM samples contain repeating structures that are similar or identical to each other.
This presents enormous redundancy in conventional data acquisition methods, 
and consequently an opportunity to design imaging systems with sparse/low-resolution samples.

In this paper, we introduce a novel computational EM imaging method based on multi-resolution data fusion (MDF) that combines innovative data acquisition with novel algorithmic techniques to dramatically improve the resolution/volume/speed trade-off as compared to traditional EM imaging methods. 
Measuring fewer samples also leads to lower dosage and less sample damage, which is crucial for imaging modalities like cryo-EM.
The key to our approach is the acquisition of sparse measurements at both low and high resolutions from which a full high-resolution image is formed.
The speedup in data acquisition results from the fact that we sparsely sample the bulk of the material while densely sampling only a small-but-representative fraction of the material -- instead of making dense high-resolution measurements across the whole material.
The high-resolution measurements are then used to create a material-specific probabilistic model in the form of a patch-library that is used to dramatically increase resolution. 

Our method depends on a Bayesian framework for the fusion of low- and high-resolution data without the need for training.
The key step in our solution is to use our library-based non-local means (LB-NLM) denoising filter as a prior model within the P\&P framework 
\cite{venkatakrishnan2013school, sreehari2016TCI} -- to produce visually true textures and edge features in image interpolations, while significantly reducing mean squared error and artifacts such as ``jaggies''. 

The P\&P framework is based on the alternating direction method of multipliers 
(ADMM) \cite{gabay1976dual, eckstein1992douglas, boyd2011distributed} and decouples the forward model  
and the prior terms in the maximum \textit{a} posteriori cost function. 
This results in an algorithm that involves repeated application of two steps: 
an inversion step only dependent on the forward model, and a denoising step only dependent on the image prior model.
The P\&P takes ADMM one step further by replacing the prior model optimization by a denoising operator of choice.

We present super-resolution and sparse interpolation results on real microscope images demonstrating data acquisition speedups of 15x to 228x on various datasets, while substantially maintaining image quality. 
The speedup factors are calculated as the ratio of the number of reconstructed pixels to the number of measured pixels.
We compare our results them with cubic interpolation and Technion's single-image super-resolution (SISR) algorithm \cite{zeyde2010single}.

\section{Related Work}
\label{section:related_work}

Image interpolation and super-resolution have been widely studied to enable high-quality imaging with fewer measurements, leading to faster and cheaper data acquisition.
Spurred by the success of denoising filters like non-local means (NLM) \cite{buades2005review, chan2014monte} in exploiting non-local redundancies, 
there have been several efforts to solve the sparse image interpolation problem using patch-based models \cite{eladHarAna05, dong2013sparse, carinSparseInterp14, sapiroGMMTIP15}.
Dictionary learning \cite{zeyde2010single, yang2010image, yang2008image} and example-based methods \cite{freeman2002example} have also been proposed for achieving super-resolution from low-resolution measurements. 
Zhang et al. \cite{zhang2012single} proposed a steering kernel regression framework \cite{takeda2007kernel} to use non-local means to achieve super-resolution.
Atkins et al. \cite{atkins1999tree} proposed tree-based resolution synthesis using a regression tree as a piece-wise linear approximation to the conditional mean of the high-resolution image given the low-resolution image.
More recently, Timofte et al. \cite{timofte2015seven} discussed the use of libraries of structures and image self-similarity to improve super-resolution, but these methods involved extensive training.
Another training-based approach to super-resolution was proposed by Perez-Pellitero et al. \cite{perez2016psyco} where they derive a regression-based manifold mapping between low- and high-resolution images.

Apart from these specific solutions for image interpolation and super-resolution, Sreehari et al. \cite{sreehari2016TCI} proposed a generic Bayesian framework called ``plug-and-play'' priors (P\&P) for incorporating modern denoising algorithms as prior models in a variety of inverse problems such as sparse image interpolation.
In this spirit, Brifman et al. \cite{brifman2016turning} have adopted the P\&P framework to use sparse-coding and dictionary-learning-based denoisers for achieving super-resolution.
In any case, no method that we know of currently exists to fuse sparse/low-resolution data with dense/high-resolution measurements to form full high-resolution images without training.

\vspace{-3mm}

\section{Method}
\label{section:method}

We present a new computational imaging method for fast acquisition of high-resolution EM images. 
Our proposed acquisition method scans the image at low resolution while simultaneously collecting a small-but-representative portion of the sample at a higher resolution. This is central to achieving the data acquisition speed-up. 
Then, we fuse the low- and high-resolution images to synthesize high resolution over the full field-of-view (FoV). In this section we describe the algorithm used to fuse the low- and high-resolution images. 
We construct a high-quality patch library containing typical textures and edge features from the high-resolution portion. 
In order to use such a library as an image model, in Sec.~\ref{section:LBNLM} we construct a version of the non-local means denoiser that uses patches from the high-resolution library.
We then use our library-based non-local means (LB-NLM) denoiser as an image/prior model within the P\&P framework (see Sec.~\ref{section:PNP}).
Using the P\&P framework, we super-resolve the low-resolution image acquired over large areas of the sample through Bayesian inversion, creating a high-resolution image over the full FoV.

\subsection{The Plug-and-Play (P\&P) Framework}
\label{section:PNP}

Let $x \in \mathbb{R}^N$ be an unknown image with a prior distribution $p(x)$,
and let $y \in \mathbb{R}^M$ be the sparsely subsampled image
with conditional distribution $p(y|x)$, which is often referred to as the forward model for the sampling system.
Then the MAP estimate of the image $x$ conditioned on the sparse/low-resolution image $y$ is given by
\begin{eqnarray}
\label{eq:MAPEst}
\hat{x}_{MAP} = \argmin_{x \in \mathbb{R}^N} \{ l(x) + \beta s(x)\},
\end{eqnarray}
where $l( x) = -\log p(y|x)$, $\beta s(x) = -\log p(x)$,
and $\beta > 0$ is a parameter used to control the regularization in the MAP reconstruction.
In order to use flexible prior models to solve ill-posed inverse problems, it is helpful to separate the forward model inversion from the prior model.
This can be effectively achieved through the application of ADMM.
The first step in applying ADMM for solving equation~\eqref{eq:MAPEst} is to split the variable $x$, resulting in an equivalent expression for the MAP estimate given by
\begin{eqnarray}
\label{eq:MAPEst_reformed}
(\hat{x}, \hat{v}) = \arg \mathop{\min_{x, v \in \mathbb{R}^N }}_{ x=v } \{l(x) + \beta s(v)\} \ .
\end{eqnarray}
This constrained optimization problem has an associated scaled-form augmented Lagrangian \cite[Section 3.1.1]{boyd2011distributed},  
\begin{eqnarray}
\label{eq:Lagrangian}
L_{\lambda}(x, v; u) = l(x) + \beta s(v) + \frac{1}{2\sigma_{\lambda}^2}\|x - v + u\|_2^2 - \frac{\|u\|^2_2}{2 \sigma_\lambda^2},
\end{eqnarray}
where $\sigma_\lambda$ is the augmented Lagrangian parameter.

The ADMM algorithm consists of iteratively minimizing the augmented Lagrangian with respect to the split variables, $x$ and $v$, followed by updating the dual variable, $u$, that drives the solution toward the constraint, $x=v$.
\begin{eqnarray}
\hat{x} &\leftarrow& \arg \min_{x\in \mathbb{R}^N } L_{\lambda}(x, \hat{v}; u) \\
\hat{v} &\leftarrow& \arg \min_{v\in \mathbb{R}^N } L_{\lambda}(\hat{x}, v ; u) \\
u &\leftarrow& u + (\hat{x} - \hat{v}) \ ,
\end{eqnarray}
where $\hat{v}$ can be initialized to a baseline reconstruction and $u$ is initialized to zero. 
By letting $\tilde{x} = v - u$, and $\tilde{v} = x + u$, 
we can define two operators that help solve  Eqs.~(4) and (5).
The first is an inversion operator $F$ defined by
\begin{eqnarray}
\label{eq:F}
F(\tilde x; \sigma_{\lambda}) = \argmin_{x \in \mathbb{R}^N } \left\{ l(x) + \frac{\|x - \tilde x\|_2^2}{2\sigma_{\lambda}^2} \right\} \ ;
\end{eqnarray}
and the second is a denoising operator $H$ given by
\begin{eqnarray}
\label{eq:H}
H(\tilde{v}; \sigma_n) = \argmin_{v \in \mathbb{R}^N} \left\{ \displaystyle\frac{\|\tilde{v}-v\|_2^2}{2\sigma_n^2} + s(v) \right\} \ ;
\end{eqnarray}
where $\sigma_n = \sqrt{\beta} \sigma_{\lambda}$ has the interpretation of being
the assumed noise standard deviation in the denoising operator.
Moreover, $H$ is the proximal mapping for the proper, closed, and convex function $s:\mathbb{R}^N \rightarrow \mathbb{R}\cup \{ +\infty \}$.

\begin{figure}[h!]
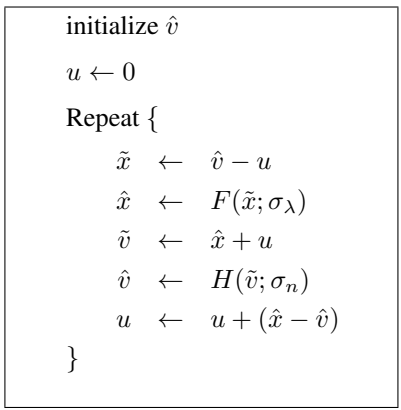

\centerline{\fbox{\parbox{5 cm}{
\begin{itemize}
\setlength\itemsep{0.5mm}
\item[] initialize $\hat{v}$ 
\item[] $u \leftarrow 0$ 
\item[] Repeat $\{$
\vspace{-2mm}
\begin{eqnarray*}
\tilde x &\leftarrow& \hat v - u \\
\hat{x} &\leftarrow& F(\tilde{x}; \sigma_{\lambda}) \\
\tilde{v} &\leftarrow& \hat{x} + u \\
\hat{v} &\leftarrow& H(\tilde{v}; \sigma_n) \\
u &\leftarrow& u + (\hat{x} - \hat{v}) \\
\end{eqnarray*}
\vspace{-14mm}
\item[] $\}$
\end{itemize}
}}}
\caption{Plug-and-play algorithm for implementation of a general forward model $F( \tilde{x} ; \sigma_\lambda )$,
and a prior model specified by the denoising operator in $H( \tilde{v} ; \sigma_n )$.} 
\label{fig:PlugAndPlayAlgorithm}
\end{figure}

Using these two operators, we can easily derive the plug-and-play algorithm
shown in Fig.~\ref{fig:PlugAndPlayAlgorithm}.

In theory, the value of $\sigma_{\lambda}$ does not affect the solution computed by the P\&P algorithm;
however, in practice a poor choice may slow convergence.
Sreehari et al. \cite{sreehari2016TCI} suggest setting the value of $\sigma_{\lambda}$ based on the amount of variation in a baseline reconstruction\footnote{In our experiments, the baseline reconstructions are Shepard or cubic interpolations.}, 
\vspace{-3mm}
\begin{equation}
\label{eq:sigma_lambda}
\sigma_\lambda^2 \approx \frac{1}{N} \sum_{i=0}^{N-1} \mathrm{var} [ x_i | y ] \ .
\end{equation}

\vspace{-5mm}

\subsection{Library-Based Non-Local Means}
\label{section:LBNLM}

Access to high-quality patches can increase the effectiveness of non-local means (NLM) as a prior model.
However, in conventional implementations of NLM, we are limited by the degraded patches present in the image. 
To remove this limitation, in this section, we outline a library-based non-local means (LB-NLM) algorithm.

Our solution is to use an external library of densely-sampled/high-resolution patches that contain typical textures and edge features.
We then compute the weighted mean of the center pixels of these patches.
Computation of the weighted mean is identical to standard non-local means, only the patches do not come from the image we are denoising. The weight normalization is also identical to standard NLM weight normalization, and is therefore simpler than the symmetric normalization (see \cite{sreehari2016TCI}).

If we denote the $N_p\times N_p$ patch centered at position $s$ by $P_s$, and the $r$-th patch of the library by $L_r$, we can compute the LB-NLM weights as
\begin{eqnarray}
w_{s,r} &\gets& \exp{\left\{ \displaystyle \frac{-\|P_s - L_r\|_2^2}{2 N_p^2 \sigma_n^2}\right\}} \\[10pt]
w_{s, r} &\gets& \frac{ w_{s, r} }{ \displaystyle \sum_{r=1}^{N_l} w_{s, r} },
\end{eqnarray}
where $N_l$ is the number of patches in the library.

\noindent Further, the LB-NLM filtering is given by,
\vspace{-1mm}
\begin{equation}
\hat{v}_s = \displaystyle \sum_{r=1}^{N_l} w_{s, r} z_r,
\end{equation}
where $\hat{v}_s$ is the denoised pixel at position $s$, and $z_r$ is the center pixel of library patch $L_r$.

We use the LB-NLM denoiser as an implicit prior model within the P\&P algorithm.
When the patch library contains high-resolution, typical textures and edge features, 
using LB-NLM as a prior model enables reproduction of visually-accurate textures and edges in the reconstructed image. 

Denoising sparse/low-resolution images with the aid of dense/high-resolution patches is, in fact, how we implement the multi-resolution data fusion in every step of the P\&P algorithm.

\subsection{Super-Resolution Forward Model}
\label{section:SR}

In this section, we formulate the explicit form of the inversion operator, $F( x, \sigma_\lambda )$,
for the application of super-resolution.
Our objective will be to interpolate an image $y\in \mathbb{R}^M$ to its higher resolution version, $x\in \mathbb{R}^N$, where $M \ll N$.
The forward model for this problem is given by
\vspace{-2mm}
\begin{equation}
\label{eq:super_resolution}
y = Ax + \epsilon \ ,
\end{equation}
\vspace{-1mm}
where $A\in \mathbb{R}^M \times \mathbb{R}^N$ represents the point-spread function (PSF) of the electron microscope.
For super-resolution by a factor of $L$, the PSF could be approximated by averaging the values of $L^2$ neighborhood pixels of every $L^2$-th pixel in $x$.
Furthermore, $\epsilon$ is an $M$-dimensional vector 
of i.i.d.\ Gaussian random variables with mean zero and variance $\sigma_w^2$.

We can write the negative log likelihood function as
\vspace{-1mm}
\begin{equation}
l(x) = \frac{ 1 }{ 2 \sigma_w^2 } \| y - A x \|^2 + \frac{ M }{ 2 } \log \left( 2 \pi \sigma_w^2 \right).
\end{equation}
In order to enforce positivity, 
we also modify the negative likelihood function by setting $l(x) = +\infty$ for $x<0$.
Using Eq.~(\ref{eq:F}), the interpolation inversion operator is then given by
\vspace{-1mm}
$$
F( \tilde{x} ; \sigma_{\lambda}) = \argmin_{x \geq 0} \left\{ \frac{1}{2 \sigma_w^2} \|y-Ax\|_2^2 + \frac{1}{2\sigma_{\lambda}^2} \|x - \tilde x\|_2^2 \right\}.
$$

If we let $\sigma_w^2 = 0$, then the inversion operator for super-resolution by a factor of $L$ reduces to the following simple expression
\vspace{-1mm}
\begin{equation}
F(\tilde x; \sigma_{\lambda}) = \left[ \tilde{x} + A^t\left(y - \frac{1}{L^2} A \tilde{x}\right)\right]_+,
\end{equation}
\vspace{-1mm}
where $[\cdot]_+$ enforces positivity.

\subsection{Sparse Interpolation Forward Model}
\label{section:Interpolation}

In this section, we formulate the explicit form of the inversion operator, $F( x, \sigma_\lambda )$,
for the application of sparse interpolation.
More specifically, our objective will be to recover an image $x\in \mathbb{R}^N$
from a noisy and sparsely subsampled version denoted by $y\in \mathbb{R}^M$
where $M \ll N$.
More formally, the forward model for this problem is given by
\vspace{-1mm}
\begin{equation}
\label{eq:sparse_interpolation}
\vspace{-1mm}
y = Ax + \epsilon \ ,
\vspace{-1mm}
\end{equation}
where the sparse sampling matrix, $A\in \mathbb{R}^M \times \mathbb{R}^N$.
Each entry $A_{i,j}$ is either 1 or 0 depending on if the $j^{th}$ pixel is taken
as the $i^{th}$ measurement.
Also, each row of $A$ has exactly one non-zero entry,
and each column of $A$ may either be empty or have one non-zero entry.
Furthermore, $\epsilon$ is an $M$-dimensional vector 
of i.i.d.\ Gaussian random variables with mean zero and variance $\sigma_w^2$.
The log likelihood has the exact form as the sparse interpolation case.

Like in the super-resolution case, if we let $\sigma_w^2 = 0$, then the inversion operator, $F$, reduces to the following explicit expression
\vspace{-1mm}
\begin{equation}
 F(\tilde x; \sigma_{\lambda}) = \left[ \tilde{x} + A^t\left(y - A \tilde{x}\right)\right]_+.
\end{equation}
where $[\cdot]_+$ enforces positivity, 
forcing the interpolation to take on the measured
values at the sample points.

\section{Results}
\label{section:Results}

In this section, we demonstrate the value of our MDF algorithm on three different datasets, all acquired on real electron microscopes. 
We compare our super-resolution reconstructions against state-of-the-art super-resolution algorithms like the single-image super-resolution (SISR) algorithm from Technion \cite{zeyde2010single}, 
and show that fusing multi-resolution data is a powerful framework to form high-quality high-resolution images -- at a fraction of the cost, time, and beam dosage compared to conventional methods.

\subsection{Datasets}

The strength of our experiments lies in the fact that all our test datasets were acquired on real electron microscopes.
As opposed to a post-processing technique, our MDF algorithm is a computational electron microscopy method that combines innovative data acquisition with novel reconstruction algorithms.

High Resolution Transmission Electron Microscopy (HR-TEM) of gold nanorods was performed on an aberration-corrected FEI Titan operating at 300 kV \cite{park2013growth}. 
Images were recorded on a 2k by 2k Gatan Ultrascan CCD at electron optical magnifications ranging from 380 kX to 640kX.
The gold atoms image (Fig.~\ref{fig:gold-atoms}) is a magnified region of edges of these gold nanorods. 

Scanning Electron Microscopy (SEM) of a surface crack in the shell of the mollusk, Hinea brasiliana \cite{deheyn2010bioluminescent} (Fig.~\ref{fig:crystal_crack}) was done on a FEI XL30 at 5 kV accelerating voltage. 

For the sparse interpolation experiment, we use a TEM image of iridovirus assemblies \cite{juhl2006assembly} (Fig.~\ref{fig:iridovirus}), which were cast onto amorphous carbon support films and imaged using a Phillips CM200 Transmission Electron Microscope. 

Note that the Hinea brasiliana image was acquired natively on the FEI Titan microscope at two different resolutions under similar imaging conditions. 
The structure in Fig.~\ref{fig:crystal_crack}(b) and~(c) were acquired at 4$\times$ resolution compared to the structure in Fig.~\ref{fig:crystal_crack}(d) and~(h).
The gold atoms image, on the other hand, was acquired at a single resolution (Fig.~\ref{fig:gold-atoms}(b)). 
We artificially reduced the resolution by 4x, 8x, and 16x 
(see Figs.~\ref{fig:gold-atoms}(c), \ref{fig:gold-atoms}(g), and \ref{fig:gold-atoms}(k), respectively).

\subsection{Experiments}

\begin{figure*}[htbp!]
\centering
\vspace{3mm}
\includegraphics[width=0.235\textwidth]{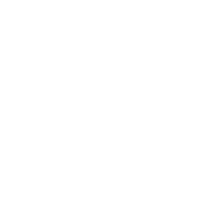}
~
\subfigure[A portion of the library image]{%
\label{fig:high_res_gold_atoms}
\includegraphics[width=0.235\textwidth]{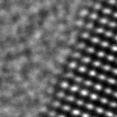}}
~
\subfigure[Ground truth]{%
\includegraphics[width=0.235\textwidth]{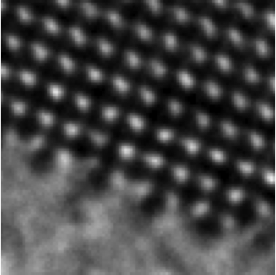}}
~
\includegraphics[width=0.235\textwidth]{figs/blank}
~

\subfigure[Low-resolution input]{%
\includegraphics[width=0.235\textwidth]{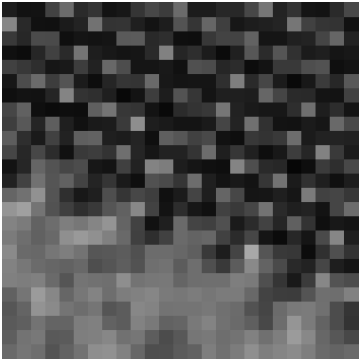}}
~
\subfigure[4x cubic interpolation]{%
\includegraphics[width=0.235\textwidth]{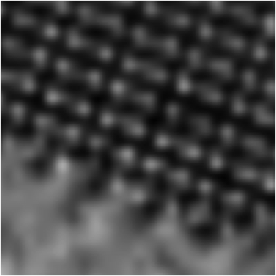}}
~
\subfigure[4x SISR]{%
\includegraphics[width=0.235\textwidth]{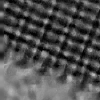}}
~
\subfigure[4x MDF (proposed algorithm)]{%
\includegraphics[width=0.235\textwidth]{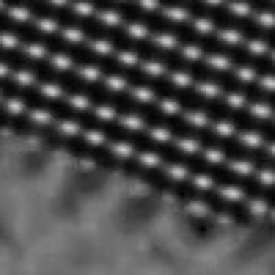}}
~

\subfigure[Low-resolution input]{%
\includegraphics[width=0.235\textwidth]{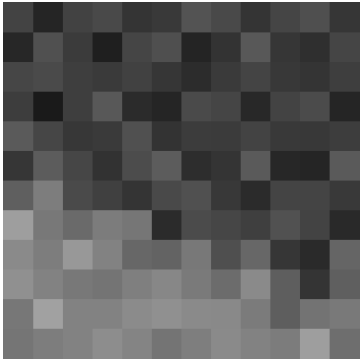}}
~
\subfigure[8x cubic interpolation]{%
\includegraphics[width=0.235\textwidth]{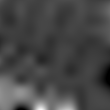}}
~
\subfigure[8x SISR]{%
\includegraphics[width=0.235\textwidth]{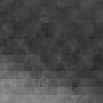}}
~
\subfigure[8x MDF (proposed algorithm)]{%
\includegraphics[width=0.235\textwidth]{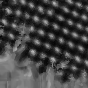}}
~

\subfigure[Low-resolution input]{%
\includegraphics[width=0.235\textwidth]{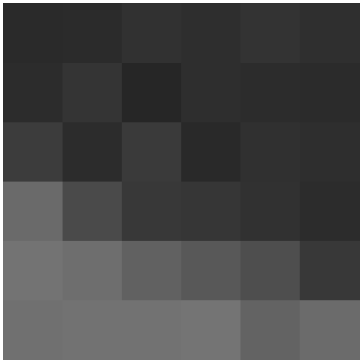}}
~
\subfigure[16x cubic interpolation]{%
\includegraphics[width=0.235\textwidth]{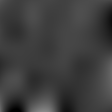}}
~
\subfigure[16x SISR]{%
\includegraphics[width=0.235\textwidth]{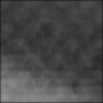}}
~
\subfigure[16x MDF (proposed algorithm)]{%
\includegraphics[width=0.235\textwidth]{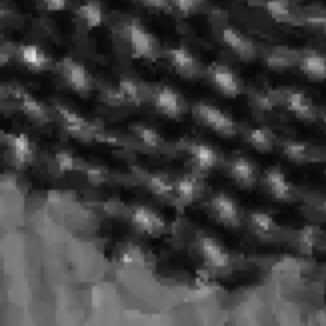}}

\caption{4x, 8x, and 16x super-resolution of an atomic-resolution TEM image of gold atoms on a carbon substrate -- zoomed in to show details. Subfigures (c), (g), and (k) were obtained by artificially reducing the resolution of the ground truth image (b), by 4x, 8x, and 16x, respectively. These low-resolution images have been rendered by pixel replication to match the size of the ground truth image. Our MDF algorithm resolves the structure of the gold atoms in all of the cases (see (f), (j), and (n)), even when the other methods struggle (see (h), (i), (l), and (m)).}
\label{fig:gold-atoms}
\end{figure*}

\begin{figure*}[htpb!]
\centering

\includegraphics[width=0.1\textwidth]{figs/blank}
~
\subfigure[A portion of the library image]{%
\label{fig:high_res_crystal}
\includegraphics[width=0.235\textwidth]{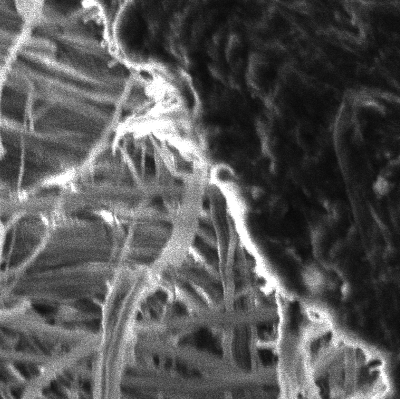}}
~
\subfigure[Ground truth]{%
\includegraphics[width=0.235\textwidth]{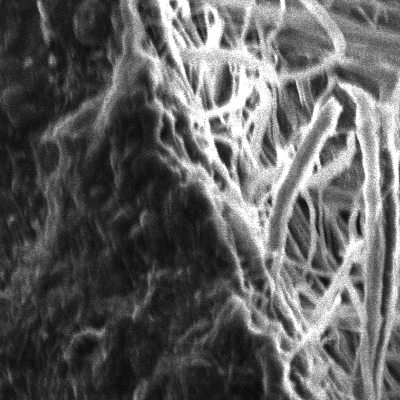}}
~
\subfigure[Ground truth (zoomed)]{%
\includegraphics[width=0.235\textwidth]{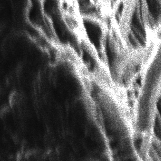}}
~
\includegraphics[width=0.1\textwidth]{figs/blank}
~
\subfigure[Low-resolution input]{%
\includegraphics[width=0.235\textwidth]{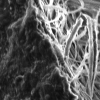}}
~
\subfigure[4x cubic interpolation]{%
\includegraphics[width=0.235\textwidth]{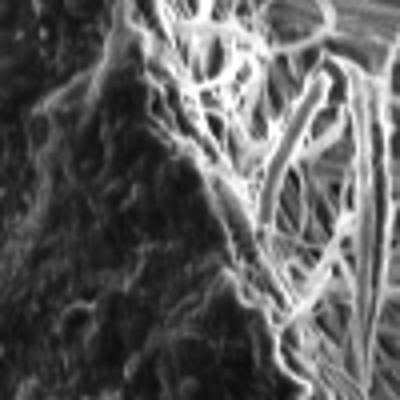}}
~
\subfigure[4x SISR]{%
\includegraphics[width=0.235\textwidth]{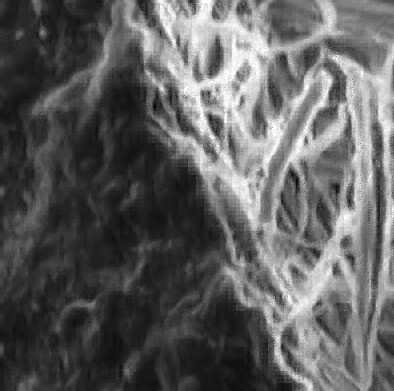}}
~
\subfigure[4x MDF (proposed algorithm)]{%
\frame{\includegraphics[width=0.235\textwidth]{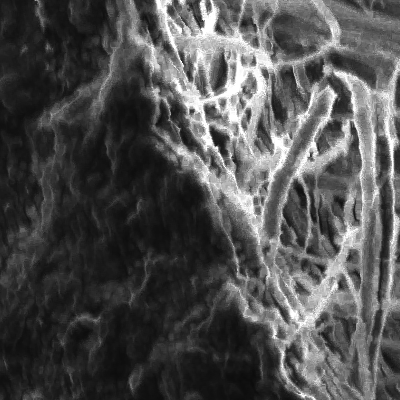}}}
~
\subfigure[Low-resolution input (zoomed)]{%
\includegraphics[width=0.235\textwidth]{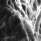}}
~
\subfigure[4x cubic interpolation (zoomed)]{%
\includegraphics[width=0.235\textwidth]{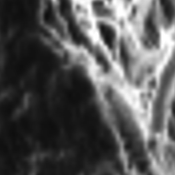}}
~
\subfigure[4x SISR (zoomed)]{%
\includegraphics[width=0.235\textwidth]{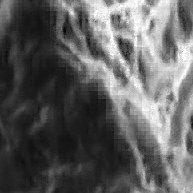}}
~
\subfigure[4x MDF (zoomed)]{%
\includegraphics[width=0.235\textwidth]{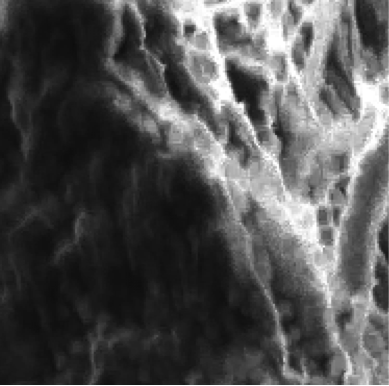}}

\caption{4x super-resolution of an SEM image of surface crack in the shell of the marine mollusk \textit{Hinea brasiliana} -- zoomed in to show details. The SEM image in (b) was acquired at 4$\times$ resolution compared to the image in (d), but under similar imaging conditions. Our MDF results in (g) and (k) show the intricate textures and clear edge features that are visually consistent with the ground truth (b). Subfigures (i) and (j) are zoomed in from (e) and (f) respectively. In (i), we see that the cubic interpolation leads to an overly smoothed interpolation, while we see in (j) that the SISR algorithm produces block artifacts.}
\label{fig:crystal_crack}
\end{figure*}

\begin{figure*}[htbp!]
\centering
\includegraphics[width=0.235\textwidth]{figs/blank}
\subfigure[Library image]{%
\label{fig:iridovirus_library}
\includegraphics[width=0.235\textwidth]{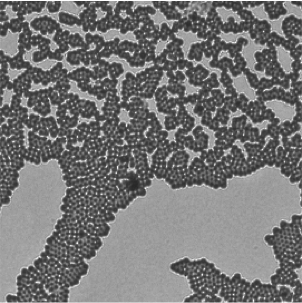}}
~
\subfigure[Ground truth]{%
\includegraphics[width=0.235\textwidth]{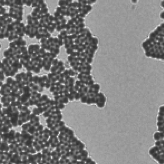}}
~
{%
\includegraphics[width=0.235\textwidth]{figs/blank}}
\subfigure[$5\%$ sampling of (b)]{%
\includegraphics[width=0.235\textwidth]{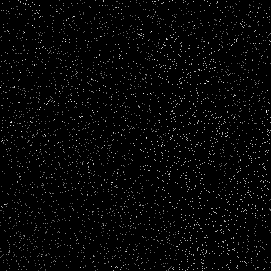}}
~
\subfigure[Shepard interpolation]{%
\includegraphics[width=0.235\textwidth]{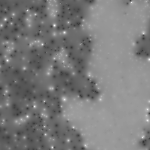}}
~
\subfigure[DSG-NLM interpolation]{%
\includegraphics[width=0.235\textwidth]{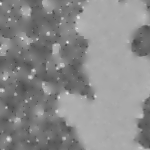}}
~
\subfigure[MDF interpolation (proposed algorithm)]{%
\includegraphics[width=0.235\textwidth]{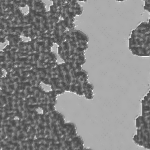}}

\caption{Interpolation of a TEM image of iridovirus assemblies on a carbon substrate; (c) shows the 5\% randomly sampled image that serves as the input to all the interpolation methods; (f) shows the texture synthesized by our MDF algorithm, which is visually more accurate compared to that generated by DSG-NLM in (e) and Shepard's interpolation in (d). In addition, the MDF interpolation results in the least mean squared error of all the methods.}
\label{fig:iridovirus}
\end{figure*}

We apply our MDF algorithm to achieve 4x, 8x, and 16x resolution synthesis results on the atomic-resolution TEM image of gold atoms (Fig.~\ref{fig:gold-atoms}), 4x resolution synthesis on the SEM image of Hinea brasiliana (Fig.~\ref{fig:crystal_crack}), 
and 5\% sparse image interpolation on the TEM image of iridovirus assemblies (Fig.~\ref{fig:iridovirus}).
We show that a variety of denoising algorithms such as doubly-stochastic gradient non-local means (DSG-NLM) \cite{sreehari2016TCI} and library-based non-local means (LB-NLM) can be plugged in as prior models within the P\&P framework to enable data acquisition speedups in EM.

In the super-resolution experiments, we compared the results of our MDF algorithm against cubic interpolation and a state-of-the-art single-image super-resolution (SISR) algorithm from Technion \cite{zeyde2010single}.
The SISR algorithm trains on high-resolution images to learn the microscope model.
To enable fair comparisons, we used the same high-resolution images to train the SISR algorithm as the ones used to create the patch library for our MDF algorithm.
Our MDF algorithm runs significantly faster than the SISR algorithm because the SISR algorithm must be trained specifically for the datasets we use\footnote{Running SISR with their pre-trained dictionaries resulted in higher RMSE.}. 
As an example, the MDF algorithm takes about 4 minutes to run to convergence on the gold atoms dataset presented here, while the SISR algorithm takes $\sim$6 hours to train and about a minute to run. 

In the sparse image interpolation experiment, we compared the result of our MDF algorithm against DSG-NLM and Shepard interpolation.\footnote{In the sparse interpolation experiment, we could not compare against the SISR algorithm since it is not designed for sparse interpolation / inpainting problems.}

To implement the LB-NLM, we built the patch-library by extracting patches from the ``library images'' (see Figs.~\ref{fig:high_res_gold_atoms}, \ref{fig:high_res_crystal}, and \ref{fig:iridovirus_library}).
We did not add noise to the measurements, and therefore set $\sigma_w = 0$. 
As stated earlier, the converged results of P\&P do not depend on the choice of $\sigma_{\lambda}^2$;
however, using the described procedure, we set $\sigma_{\lambda}^2 = 64$ for sparse interpolation, $\sigma_{\lambda}^2 = 55$ for super-resolution of hinea brasiliana, and $\sigma_{\lambda}^2 = 72$ for super-resolution of gold atoms.
The only free parameter for P\&P reconstruction is then the unit-less parameter 
$\beta$,
which we selected to minimize the mean squared error of the reconstruction compared to the ground truth (see Table~\ref{table:parameters}).
For all experiments, we present the P\&P residual convergence resulting from using different priors (see Tables~\ref{table:primal_residues}).
The normalized residue \cite[p. 18]{boyd2011distributed}, $r^{(k)}$ is given by
\begin{eqnarray}
r^{(k)} = \displaystyle\frac{\| \hat{x}^{(k)} - \hat{v}^{(k)} \|_2}{\| \hat{x}^{(\infty)} \|_2} \,
\end{eqnarray}
where $\hat{x}^{(k)}$ and $\hat{v}^{(k)}$ are the values of $\hat{x}$ and $\hat{v}$ respectively after the $k$-th iteration of the P\&P algorithm, respectively, and $\hat{x}^{(\infty)}$ the final value of the reconstruction, $\hat{x}$.


\begin{table}[h]
\begin{center}
\begin{tabular}{|l|ccc|}
\hline
\multicolumn{4}{ |c| }{\textbf{RMS Error of the Super-Resolution Experiments}} \\
\hline\hline
\textbf{Dataset / } & \textbf{Cubic} & \textbf{SISR} & \textbf{MDF} \\
\textbf{resolution synthesized} & & & \\
\hline\hline
Hinea brasiliana / 4x & 10.33\% & 6.14\% & \textbf{4.10\%} \\
\hline
Gold atoms / 4x & 16.12\% & 11.73\% & \textbf{7.87\%} \\
\hline
Gold atoms / 8x & 39.36\% & 32.58\% & \textbf{12.65\%} \\
\hline
Gold atoms / 16x & 45.71\% & 38.11\% & \textbf{20.89\%} \\
\hline
\end{tabular}
\end{center}
\vspace{-2mm}
\caption{RMSE between the super-resolution interpolated image and the high-resolution ground truth.}
\label{table:super-resolution_error}
\end{table}

\vspace{-2mm}

\begin{table}[H]
\begin{center}
\begin{tabular}{|l|ccc|}
\hline
\multicolumn{4}{ |c| }{\textbf{RMS Error of the Sparse Interpolation Experiment}} \\
\hline\hline
\textbf{Experiment} & \textbf{Shepard} & \textbf{DSG-NLM} & \textbf{MDF} \\
\hline\hline
Sparse interpolation  & 13.10\% & 10.64\% & \textbf{9.71\%} \\
\hline
\end{tabular}
\end{center}
\vspace{-2mm}
\caption{RMSE between the interpolated image and the ground truth.}
\label{table:interpolation_error}
\end{table}

\begin{table}[h]
\begin{center}
\begin{tabular}{|l|cc|}
\hline
\multicolumn{3}{ |c| }{\textbf{Convergence Error of P\&P}} \\
\hline\hline
\textbf{Experiment} & \textbf{DSG-NLM} & \textbf{MDF} \\
\hline\hline
Sparse interpolation & \textbf{$4.66\times 10^{-8} \%$} & \textbf{$8.98\times 10^{-9} \%$} \\
\hline
Super-resolution & -- & \textbf{$5.91\times 10^{-7} \%$} \\
(hinea brasiliana) & & \\
\hline
Super-resolution & -- & \textbf{$6.09\times 10^{-7} \%$} \\
(gold atoms) & & \\
\hline
\end{tabular}
\end{center}
\vspace{-2mm}
\caption{Convergence error of the P\&P algorithm}
\label{table:primal_residues}
\end{table}

\begin{table}[htpb!]
\begin{center}
\begin{tabular}{|l|cc|}
\hline
\multicolumn{3}{ |c| }{\textbf{Regularization Parameter, $\beta$}} \\
\hline\hline
\textbf{Experiment} & \textbf{DSG-NLM} & \textbf{MDF} \\
\hline\hline
Sparse interpolation  & 0.39 & 0.42 \\
\hline
Super-resolution & -- & 0.51 \\
(hinea brasiliana) & & \\
\hline
Super-resolution & -- & 0.36 \\
(gold atoms) & & \\
\hline
\end{tabular}
\end{center}
\vspace{-2mm}
\caption{Regularization parameter, $\beta$}
\label{table:parameters}
\end{table}

Our MDF reconstructions result in 15x to 228x data acquisition speedup across our datasets. 
The speedup factors are computed as the ratio of the number of pixels in the reconstructed image to the number of pixels measured. In addition to specifying the speedup factors, we specify the size of our high-resolution library relative to the size of the low-resolution image acquired through the quantity,
\begin{equation}
\rho = \frac{\text{\# of high-resolution pixels}}{\text{\# of total pixels}}.
\end{equation}
Table~\ref{table:speedup_factors} gives the value of $\rho$ and the speedup factors for each of our experiments.
\begin{table}[htpb!]
\begin{center}
\begin{tabular}{|l|cc|}
\hline
\multicolumn{3}{ |c| }{\textbf{Data Acquisition Speedup}} \\
\hline\hline
\textbf{Experiment} & \textbf{$\rho$} & \textbf{Speedup factor} \\
\hline\hline
Super-resolution & 6.11\% & 15x \\
(hinea brasiliana / 4x) & & \\
\hline
Super-resolution & 0.76\% & 15.87x \\
(gold atoms / 4x) & & \\
\hline
Super-resolution & 2.96\% & 62.1x \\
(gold atoms / 8x) & & \\
\hline
Super-resolution & 10.88\% & 228.14x \\
(gold atoms / 16x) & & \\
\hline
Sparse interpolation  & 8.74\% & 18.25x \\
(iridovirus)  & &  \\
\hline
\end{tabular}
\end{center}
\vspace{-2mm}
\caption{Data acquisition speedup achieved by our MDF algorithm. The ratio of the high-resolution pixels measured to the low-resolution pixels measured is given by $\rho$.}
\label{table:speedup_factors}
\end{table}

Importantly, the MDF reconstructions are clearer than Shepard and cubic interpolations, despite the high speedup factors achieved. 
Also, MDF results in the least mean squared error (see Tables~\ref{table:super-resolution_error} and \ref{table:interpolation_error}).
The MDF algorithm also substantially reduces jaggies and produces better textures and features compared to Technion's SISR algorithm as well as cubic and Shepard interpolations.

\section{Conclusion}
\label{section:Conclusion}

Traditional EM imaging based on homogeneous raster-order scanning severely limits the volume of high-resolution data that can be collected, 
and presents a fundamental limitation to understanding many important physical processes.

In this paper, we introduced a novel computational electron microscopy imaging method based on multi-resolution data fusion (MDF) that combines innovative data acquisition with novel algorithmic techniques to dramatically improve the resolution/volume/speed/dosage trade-off as compared to traditional EM imaging methods. 
High-resolution measurements taken over a small field-of-view are used to create a material-specific probabilistic model in the form of a patch-library that is used to dramatically increase resolution. 
We introduced a Bayesian framework for the fusion of high and low-resolution data without the need for training. 
We outlined a library-based non-local means (LB-NLM) algorithm that uses typical high-resolution patches, 
and we used it as a prior model within the \textit{plug-and-play priors} framework to produce visually true textures and edge features in the interpolations, while significantly reducing mean squared error. 
Finally, we presented 4x, 8x, and 16x super-resolution and 5\% sparse image interpolation results demonstrating data acquisition speedups of 15x to 228x on various datasets, while substantially maintaining image quality. 
Our MDF algorithm greatly increases acquisition speed and reduces the electron beam dosage.
We demonstrated the power of our method on real electron microscope images and compared our results against Technion's state-of-the-art single-image super-resolution (SISR) algorithm.
In our experiments, we found our MDF algorithm to consistently produce clearer, sharper image reconstructions with visibly truer textures and lower mean squared error compared to the SISR, cubic, and Shepard interpolations.

In summary, our MDF computational EM method enables high-resolution EM data to be collected at a fraction of the time as conventional methods, while maintaining high image quality -- offering an \textit{integrated imaging} solution to a grand challenge problem in electron microscopy.

{\small
\bibliographystyle{ieee}
\bibliography{References}
}

\end{document}